\documentclass[12pt]{article}
\usepackage[a4paper, total={6in, 8in}]{geometry}

\usepackage{amsmath}
\usepackage{amsthm}

\usepackage{graphicx}
\usepackage{booktabs} 
\usepackage{graphicx}
\usepackage{subcaption} 
\usepackage{newtxtext}
\usepackage[subscriptcorrection]{newtxmath}

\usepackage{natbib}
\usepackage{hyperref}

\usepackage{cleveref}   
\graphicspath{{./art/}}

\usepackage{setspace}

\makeatletter

\begin{document}

\newtheorem{theorem}{Theorem}[section]
\newtheorem{lemma}[theorem]{Lemma}
\newtheorem{proposition}[theorem]{Proposition}
\newtheorem{corollary}[theorem]{Corollary}

\theoremstyle{definition}
\newtheorem{definition}[theorem]{Definition}
\newtheorem{assumption}[theorem]{Assumption}

\theoremstyle{remark}
\newtheorem{remark}[theorem]{Remark}


\title{Understanding Over-parametrization in Survival Models through Interpolation}

\author{
Yin Liu,~Jianwen Cai,~Didong Li\thanks{\texttt{didongli@unc.edu}}\\
Department of Biostatistics, University of North Carolina at Chapel Hill.}

\date{}

\maketitle

\setstretch{1.5}
\begin{abstract}
Modern statistical learning theory has challenged the classical bias-variance trade-off by exhibiting the \textit{double-descent} phenomenon for over-parametrized models, where test loss (or risk) decreases again after peaking near the \textit{interpolation} threshold. While a rigorous theory of \textit{double-descent} has been established for linear models, a unified characterization for deep neural networks remains an area of active research. Its manifestation in survival analysis is even less understood, as the presence of censoring and the use of likelihood-based losses make defining the \textit{interpolation} threshold mathematically non-trivial. To address this gap, we establish a generalized framework for all loss-based problems by formally defining \textit{interpolation} and \textit{finite-norm interpolation}, treating regression, classification, and survival analysis as special cases. Specifically, we rigorously prove the existence or absence of \textit{interpolation} and \textit{finite-norm interpolation} for four representative deep survival models: DeepSurv, PC-Hazard, Nnet-Survival, and N-MTLR, clarifying how likelihood-based losses and implementation choices jointly determine \textit{interpolation} feasibility. Supported by numerical experiments, our findings highlight the distinct generalization behaviors of these four models and demonstrate that over-parametrization should not be regarded as benign in the survival context. 
\end{abstract}


\section{Introduction}

\begin{figure}[t]
\vspace{-0.1in}
\centering
    \centering
    \includegraphics[width=0.49\linewidth]{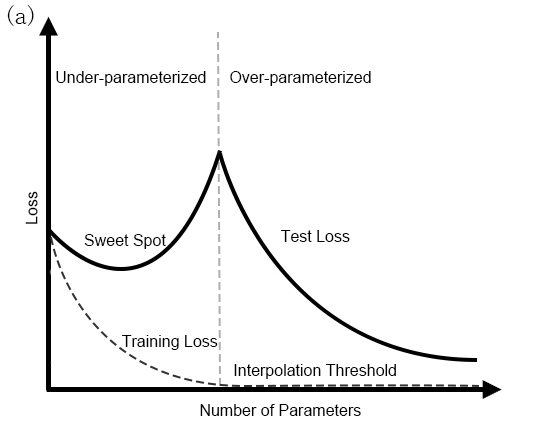}
    \includegraphics[width=0.49\linewidth]{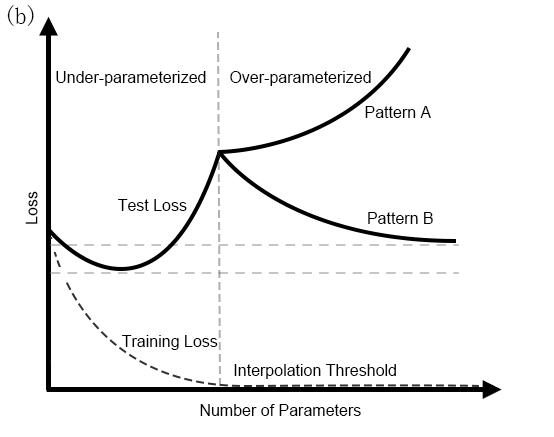}
    \label{fig:patterns}
\vspace{-0.15in}
\caption{
Conceptual illustration of the double-descent phenomenon.
(a) Classical double-descent in regression or classification, where test error decreases again beyond the interpolation threshold.
(b) Possible variants of double-descent–like behavior in survival models.
}
\vspace{-0.15in}
\label{fig:combined}
\end{figure}

For decades, the bias-variance trade-off has served as a cornerstone for understanding model generalization~\citep{hastie2009elements}. Classical learning theory suggests that as model capacity (i.e., number of parameters) increases, training loss (also referred to as the risk, error, or objective function; hereafter simply called the loss) decreases monotonically while test loss first decreases and then increases, yielding the familiar U-shaped risk curve~(\Cref{fig:combined}a). This view implies that models fitting the training data too closely will inevitably overfit and generalize poorly. In recent years, however, this belief has been challenged by a striking empirical observation known as the double-descent phenomenon~\citep{belkin2019reconciling,nakkiran2021deep}. As model complexity grows beyond the point of perfect interpolation, where the training loss achieves zero, the test loss often decreases again, forming a second descent as shown in Figure \ref{fig:combined}a. This pattern has been documented across a wide range of models, including linear and logistic regression with random features~\citep{muthukumar2021classification,hastie2022surprises,kini2020analytic}, kernel methods~\citep{nakkiran2021deep,belkin2019reconciling}, random forests~\citep{belkin2019reconciling}, and deep neural networks~\citep{nakkiran2021deep,singh2022phenomenology}. These findings have prompted a re-examination of the classical bias-variance framework and suggest that generalization in the modern, over-parameterized regime may follow a richer geometry than the traditional U-shape.

Although often discussed as a single empirical phenomenon, double-descent in fact reflects several distinct theoretical mechanisms. 
From a theoretical standpoint, it is therefore helpful to disentangle a sequence of logically related questions rather than treating double-descent as a monolithic pattern in test loss.

The first question concerns the existence of interpolation: as model capacity increases, does a well-defined interpolation threshold exist? 
This issue has been relatively well studied in linear models, random-feature models, and certain ensemble methods~\citep{hastie2022surprises,bach2024double_descent_random_projections}.
The second question asks whether test loss exhibits a second descent once interpolation is achieved. When the second descent exists, the third question concerns whether the post-interpolation test loss can fall below the minimum of the classical U-shaped risk curve. 
To date, the second and the third questions remain theoretically unresolved in general, except in highly structured settings such as minimum-norm linear regression~\citep{hastie2022surprises,bach2024double_descent_random_projections} and kernel regression (including its neural tangent kernel equivalence), where this behavior can be rigorously derived under specific assumptions on the data-generating mechanism and model misspecification~\citep{hastie2022surprises}. For more complex model families, including deep neural networks and random forests, general theoretical conclusions remain open. 

While double-descent has been studied in regression and classification, its behavior under survival analysis remains largely unknown. 
It is unclear whether survival models follow the canonical pattern shown in Figure~\ref{fig:combined}a, where the test loss decreases again beyond the interpolation threshold and falls below the initial sweet-spot level. 
Alternatively, they may exhibit other patterns illustrated in Figure~\ref{fig:combined}b, such as Pattern~A, where the loss continues to rise with increasing capacity, or Pattern~B, where it declines after interpolation but remains above the sweet-spot level. These ambiguities motivate a more careful examination of interpolation itself, including its definition and existence, as a foundational step toward understanding double-descent behavior and over-parametrization in survival analysis.

Survival analysis occupies a distinctive position in statistical learning \citep{collett2023modelling}, and two features make it fundamentally different from standard supervised settings. First, its loss functions are derived from partial or full likelihoods rather than mean-squared or cross-entropy losses
\citep{klein2003survival}. The likelihood formulation changes both the mathematical form and interpretation of the loss function. Unlike standard loss functions that are evaluated directly on observed outcomes, likelihood-based losses represent the probability of the observed event pattern under a model. This distinction makes interpolation less straightforward to define, since maximizing the likelihood involves fitting a probabilistic model rather than matching each observation exactly. 
Second, under typical model assumptions, survival neural networks do not predict directly observable quantities. Instead, they estimate latent structures such as hazard or risk functions that describe the underlying event process~\citep{kleinbaum2012survival}. Even with complete data, these targets cannot be observed explicitly, which means that ``fitting the training data exactly'' has no clear or unique meaning. This difference between predicting observables and estimating latent mechanisms adds an additional layer of abstraction to the learning process and may fundamentally alter how over-parameterization affects generalization. 
Together, these two characteristics make survival learning a compelling yet challenging setting for studying whether double-descent reflects a general property of modern learning or one that depends critically on how the loss links model predictions to the data-generating process. From a practical perspective, deep survival networks have become increasingly common in biomedical and clinical prediction~\citep{wiegrebe2024deep}, yet their generalization under heavy over-parameterization remains poorly characterized. A systematic study in this context is therefore both timely and important.

To address this gap, we first rigorously define interpolation and finite-norm interpolation for generic loss-based models, including regression, classification, and survival models as special cases. We further develop theoretical analyses on the existence of interpolation and finite-norm interpolation for four representative state-of-the-art deep survival models, and link these theoretical conditions to the emergence of double-descent behavior. Specifically, we investigate DeepSurv (Deep Neural Network for Cox Proportional Hazards;~\citealp{katzman2018deepsurv}) and PC-Hazard (Piecewise-Constant Hazard Model;~\citealp{kvamme2019time}) for continuous time,
Nnet-Survival (Neural Network-Based Discrete-Time Survival Model;~\citealp{gensheimer2019nnet}), and N-MTLR (Neural Multi-Task Logistic Regression;~\citealp{Fotso2018DeepNN}) for discrete time.

Across all four models, the training loss decreases monotonically and approaches its theoretical infimum, 
indicating that approximate interpolation is achieved. 
Yet the corresponding test-loss trajectories differ markedly. For continuous-time models (DeepSurv and PC-Hazard), the test loss follows a recognizable double-descent trend, similar to Pattern B in \Cref{fig:combined}b: 
it first decreases, rises near the interpolation threshold, and then declines again. 
In PC-Hazard, this second descent is clear but stabilizes above the classical bias-variance minimum, 
while in DeepSurv it is weaker and less distinct, suggesting only limited benefit from increasing capacity. 
For discrete-time models (Nnet-Survival and N-MTLR), the test loss shows only a brief and shallow dip immediately after the interpolation point, 
followed by a sharp and sustained increase, consistent with Pattern A in Figure~\ref{fig:combined}b. 
This narrow quasi-descent window indicates that over-parameterization yields little practical gain across all models. These results reveal that survival learning exhibits distinct generalization behavior compared with standard regression or classification settings, reflecting how model capacity and loss formulation jointly shape the loss landscape.

Our work contributes to the broader effort to connect modern machine learning practice with classical statistical principles in the presence of censoring. 
The theoretical analysis, with proofs that are model-dependent, highlights how distinct loss geometries give rise to different interpolation behaviors. All proofs are postponed to the Appendix~\ref{appendix_A}. Beyond its theoretical contribution, these findings provide practical guidance for model design and training in survival analysis. Although certain continuous-time models display a mild second descent in the over-parameterized regime, their generalization loss remains substantially higher than the classical bias-variance minimum. This indicates that over-parametrization should not be regarded as benign in practice. Overall, our study clarifies where stable interpolation ends and harmful over-parameterization begins and provides new insight into how deep survival networks behave across different capacity regimes.

\section{Background and related work}

In classical statistical learning theory, increasing model complexity typically reduces bias but increases variance beyond a certain point known as the ``sweet spot'', leading to the familiar U-shaped test loss curve that underlies the bias–variance trade-off. Specifically, the training loss monotonically decreases with model capacity, while test loss initially falls, reaches a minimum, and then rises again as overfitting occurs.

However, recent advances challenge this U-shaped view of generalization. The double-descent phenomenon, first formalized by \cite{belkin2019reconciling}, shows that as the model capacity continues to grow, the test loss decreases again beyond a certain point, known as the interpolation threshold. Although the double-descent phenomenon has been observed across many models, its emergence and shape depend crucially on the geometry and smoothness of the loss function. Different losses impose distinct structural constraints on the optimization landscape, leading to markedly different generalization dynamics around the interpolation threshold. In this sense, double-descent arises not from model capacity alone but from how the loss geometry governs the evolution of test loss. In this section, we introduce the background and summarize existing results on double-descent.

\subsection{Existing results for various loss functions}\label{subsec:loss_specific}


Consider a standard prediction setup with observations $\{(x_i,y_i)\}_{i=1}^n$, where $x_i\in\mathbb{R}^p$ and $y_i\in\mathbb{R}$. Since interpolation is defined relative to a realized dataset, namely as exact fit to the training data \citep{belkin2019reconciling}, analyses of interpolation and double descent are naturally formulated conditional on the dataset. Throughout this work, we adopt this perspective and treat the dataset as fixed unless otherwise noted. When $y_i$ is continuous, the prediction task is regression; when $y_i$ is discrete, the task is classification. Here,  we consider a generic family of models $\{f_\theta:\mathbb{R}^p\to\mathbb{R}|\theta\in\Theta\subset\mathbb{R}^d\}$ where the parameter space is $\Theta$ with dimension $d$. The interpolation property highly depends on the loss function. Table~\ref{table1} present the loss functions that have been studied in the double-descent literature.. 

\begin{table}[!h]
  \caption{Various loss functions studied in the double-descent literature.}
  \vspace{0.3cm}
  \label{table1}
  \begin{center}
    \begin{small}
      \begin{sc}
        \begin{tabular}{lcccr}
          \toprule
          Loss Name  & Function        \\
          \midrule
          Squared loss & $
             \frac{\sum_{i=1}^n\big(f_\theta(x_i) - y_i\big)^2}{n}
            $  \\
          Logistic loss  & $\frac{\sum_{i=1}^n \log\!\big(1 + \exp(-y_i f_\theta(x_i))\big)}{n}$ \\
          Cross entropy   &  $\frac{-\sum_{i=1}^n y_i \log p_\theta(x_i)+(1-y_i)\log(1-p_\theta(x_i))}{n}$\\
          Zero-one loss    & $\frac{\sum_{i=1}^n\mathbb{I}\big[\operatorname{sign}(f_\theta(x_i)) \neq y_i\big]}{n}$ \\
          Hinge loss & $\frac{\sum_{i=1}^n\max(0, 1 - y_i f_\theta(x_i))}{n}$\\
          \bottomrule
        \end{tabular}
      \end{sc}
    \end{small}
  \end{center}
  \vskip -0.1in
\end{table}

\noindent\textit{Squared loss. } For continuous outcome $y$, the squared loss is smooth, convex, and analytically tractable, making it the canonical setting for studying double-descent. \cite{belkin2019reconciling} showed in random-feature and neural-network models that when capacity first becomes sufficient to achieve zero training loss, 
the test loss peaks at this interpolation point and then declines as capacity increases further, showing a clear ``second descent''. \cite{hastie2022surprises} derived the corresponding high-dimensional risk formula: 
with aspect ratio $\gamma = d/n$, the variance explodes near $\gamma \approx 1$ but rapidly decreases for $\gamma > 1$.

\noindent\textit{Logistic loss.} For binary classification, the logistic loss originates from the maximum-likelihood principle.
 For separable data (i.e., $\exists~\theta_*$ such that $y_i f_{\theta_*}(x_i)>0,~\forall i$), driving the training loss toward zero requires $y_i f_\theta(x_i) \to \infty$; 
thus, exact interpolation cannot be achieved with parameters with bounded norm (known as finite-norm interpolation). 
Gradients $\nabla_\theta\mathcal{L}(\theta)$ therefore flatten sharply near the interpolation region, diminishing sensitivity to capacity and weakening the second descent. \cite{kini2020analytic} confirmed that, in high-dimensional linear models, logistic loss exhibits a sharper peak and almost no secondary decrease compared with square loss. 

\noindent\textit{Cross entropy.}  For binary classification, the cross entropy loss uses $p_\theta(x_i)$ as an estimate of $P(Y=1|X=x_i)$ usually determined by $f_\theta(x_i)$. Consistent behavior appears in deep networks trained with cross-entropy \citep{nakkiran2021deep, singh2022phenomenology}: 
after near-interpolation, the loss curve enters a ``saturation plateau,'' flattening or oscillating rather than descending again.

\noindent\textit{Zero-one loss.} The zero-one loss is discrete and non-differentiable, and the training loss drops immediately to zero once the data become separable. 
Belkin et al. further showed that, for otherwise identical models, the square loss yields a double-descent loss curve, 
whereas the zero-one loss produces a monotone decrease~\citep{belkin2019reconciling} .

\noindent\textit{Hinge loss.} The hinge loss leads to the maximum-margin solution in separable cases; gradients vanish outside the margin, preventing a smooth transition through the interpolation region. 
Hence, when the loss loses gradient continuity near zero, the model rarely displays a classical double-descent pattern.

\subsection{Theoretical extension}

The loss-specific observations above motivate a more unified perspective on how loss geometry influences generalization near the interpolation threshold. Muthukumar et al.~\cite{muthukumar2021classification} further demonstrated that the interpolation geometry itself depends on the choice of loss. 
In over-parameterized linear models, the minimum-$L_2$-norm interpolator under square loss and the maximum-margin solution under hinge loss coincide in form, 
yet their test loss curves differ markedly. 
This finding suggests that the existence and shape of double-descent are primarily governed by the loss geometry around zero training loss, 
not by the model family or task. 
Smooth losses enabling finite-norm interpolation (e.g., square loss) tend to produce a clear rise-and-fall pattern, 
whereas losses whose gradients flatten or become discontinuous near the margin (e.g., logistic, hinge, or zero-one) yield monotonic or weakly decaying curves.


Similar trends extend beyond linear models. 
In tree-based ensembles \citep{belkin2019reconciling}, 
double-descent-like behavior appears only when multiple complexity dimensions, such as tree depth and ensemble size, grow jointly. 
These results reinforce that the manifestation of double-descent is driven by the interplay between loss geometry and capacity growth, 
rather than by any single modeling architecture.

Overall, the appearance of double-descent hinges on how the loss behaves as the training loss approaches zero. 
When the loss remains smooth and provides continuous gradients (as in square loss), 
the test loss curve near the interpolation threshold exhibits a distinct rise-and-fall pattern. 
When gradients vanish or the loss becomes discrete (as in logistic, hinge, or zero-one formulations), 
models may still achieve or asymptotically approach interpolation, yet the loss curve typically lacks a pronounced second descent. 
Thus, double-descent fundamentally reflects the interaction between loss geometry and capacity scaling, 
providing a theoretical foundation for later analyses of likelihood-based survival loss functions.

\subsection{Loss functions in survival models}

Survival models, by contrast, have received little attention from the perspective of double-descent. Recent years have seen the emergence of neural extensions of classical survival models, which can be broadly grouped by their treatment of event time into {continuous-time} and {discrete-time} formulations. Among continuous-time approaches, models such as DeepSurv \citep{katzman2018deepsurv} extend the Cox proportional hazards framework through the partial likelihood, while PC-Hazard \citep{kvamme2019time} specifies a full likelihood by directly modeling the hazard function using piecewise constants. In the discrete-time domain, Nnet-Survival \citep{gensheimer2019nnet} represents the conditional event probability at each time step through a softmax-like structure, whereas N-MTLR \citep{Fotso2018DeepNN} models the survival function via a sequence of dependent logistic regressions across ordered intervals. Despite their differences in likelihood construction, these models share a common feature: they rely on likelihood-based optimization rather than direct loss minimization. While they have demonstrated strong empirical performance, their generalization behavior under over-parameterization, and even the definition of interpolation under such likelihood-based loss functions, remains poorly understood.

Our analysis proceeds at the level of individual loss functions rather than model families. Each survival loss function, such as those underlying DeepSurv, PC-Hazard, Nnet-Survival, and N-MTLR, imposes its own structural constraints that govern whether interpolation can exist mathematically and whether it can be attained in practice. By examining these loss-specific geometries under controlled synthetic settings, we clarify when interpolation is theoretically feasible and whether it occurs empirically, laying the groundwork for understanding double-descent in likelihood-based survival modeling.

\section{Interpolation and finite-norm interpolation}

We consider a given right-censored survival dataset
$\mathcal{D} = \{(x_i, T_i, \delta_i)\}_{i=1}^{n}$,
where $x_i \in \mathbb{R}^p$ denotes the covariate vector,
$T_i \in \mathbb{R}_+$ is the observed event or censoring time,
and $\delta_i \in \{0,1\}$ is the event indicator 
($\delta_i = 1$ for an event and $\delta_i = 0$ for censoring).
Let $\mathcal{R}_i = \{ j : T_j \ge T_i \}$ denote the risk set at time $T_i$.  In line with the perspective introduced in Section~\ref{subsec:loss_specific}, we treat the dataset $\mathcal{D}$ as fixed throughout. Consequently, all loss functions are understood to be functions of parameters conditional on the data $\mathcal{D}$, and this dependence is omitted from the notation.
Each neural survival model defines a mapping $z_\theta(\cdot)$,
parameterized by $\theta$, which outputs either a scalar log-risk score or
a vector of logits across time intervals, depending on the model.
We use $z_\theta(\cdot)$ for notational consistency,
while recognizing that how it enters the loss is model-specific.

In the context of regression or classification, interpolation has a precise and tractable meaning, i.e., the predicted outcomes exactly match the observed outcomes $f_\theta(x_i)=y_i$, achieving a zero training loss. This formulation assume that every observation $x_i$ contributes a directly observed outcome $y_i$, allowing a well-defined notion of perfect fit. 

Survival learning, however, violates this assumption. Its loss functions are likelihood-based and depend on censored event processes rather than observed outcomes. Many observations are only partially known through censoring indicators, and the model targets latent risk or hazard structures rather than realized event times. Consequently, even defining what it means to interpolate the training data is non-trivial: a perfectly specified model cannot reproduce unobserved event times, and the likelihood may not admit an exact interpolating solution in the classical sense.

This conceptual difficulty motivates our study. We provide the first systematic examination of interpolation in survival models, focusing on how this concept can be meaningfully characterized under likelihood-based losses. To formalize this notion, we adopt a unified definition of interpolation that applies consistently across all loss-based models, including but not limited to prediction and survival models: 

\begin{definition}[Interpolation]\label{def:interpolation_math}
For a given dataset $\mathcal{D}$, the interpolation threshold of a model with loss $\mathcal{L}$ is defined as
\[
d^* \;=\; \arg\min_{d \in \mathbb{N}_+} 
\Big( \inf_{\theta \in \mathbb{R}^d} \, \mathcal{L}(\theta) \Big),
\]
that is, the smallest number of parameters for which the loss \(\mathcal{L}(\theta)\) attains its infimum and cannot be further reduced by enlarging the parameter space. The model is said to admit an interpolation if $d^*<\infty$ for any dataset $\mathcal{D}$.
\end{definition}

It aligns the notion of ``perfect fit’’ in survival learning with the general interpolation framework: the point where the loss function has reached its infimum, and enlarging the model space no longer improves the fit.

While the preceding analysis establishes when interpolation is algebraically possible, it does not guarantee that such interpolation can be achieved within a bounded parameter space within $\mathbb{R}^d$.
In many over-parameterized models, the empirical loss may approach zero only as the parameter norm diverges, for instance, in logistic or softmax regression, exact interpolation requires $\|\theta\|\to\infty$ \citep{nakkiran2021deep}.
Most classical double-descent analyses identify the interpolation threshold merely as the point where zero training loss becomes attainable \citep{advani2020highdim, nakkiran2021deep}, without explicitly requiring that the interpolating solution remain bounded.
However, the sharpest and most interpretable double-descent curves have been observed in models that admit finite-norm interpolating solutions, such as minimum-norm linear regression and random-feature regression \citep{belkin2019reconciling, hastie2022surprises}.

This link suggests that finite-norm interpolation governs not only the existence of perfect fitting, 
but also the visibility of double-descent in practice. 
When a model’s loss surface admits a broad near-zero-loss plateau within finite norms, test loss can decrease again beyond the interpolation threshold, forming the characteristic second descent. 
In contrast, when approaching the infimum requires unbounded parameters, as in most likelihood-based survival loss functions~(see \Cref{sec:theory}), the plateau collapses, 
and double-descent becomes weak or absent.

Although the distinction between the mere existence of an interpolation and the existence of a finite-norm interpolation has been noted implicitly in the regression literature \citep{muthukumar2020harmless, liang2020multiple}, it has not been made explicitly and remains unexplored in the survival setup.
As a result, we reexamine the notion of finite-norm interpolation in a general framework analogous to Definition \ref{def:interpolation_math}, which determines not only whether the interpolating solution is geometrically realizable, but also whether the corresponding double-descent behavior can be sharply observed in practice.

\begin{definition}[Finite-norm interpolation]\label{def:finite-interpolation}
For any given dataset $\mathcal{D}$, the finite-norm interpolation threshold of a model with loss $\mathcal{L}$ as
\[
d^*_{\mathrm{F}} \;=\; \arg\min_{d \in \mathbb{N}_+} 
\Big( \inf_{\theta \in [-M,M]^d} \, \mathcal{L}(\theta) \Big),
\]
for some $M>0$. That is, the smallest number of parameters for which 
the loss \(\mathcal{L}(\theta)\) attains its infimum and 
cannot be further reduced by enlarging the parameter space.  The model is said to admit a finite-norm interpolation if $d^*_{\mathrm{F}}<\infty$ for any dataset $\mathcal{D}$.
\end{definition}

Throughout the remainder of this paper, all references to interpolation and finite-norm interpolation are understood according to Definitions~\ref{def:interpolation_math} and~\ref{def:finite-interpolation}. In the next section, we develop theoretical results characterizing interpolation and finite-norm interpolation properties of four representative survival models and analyze how its presence or absence shapes the resulting generalization behavior.

\section{Theoretical analysis}\label{sec:theory}
Building on the rigorous definition of interpolation and finite-norm interpolation, we next provide theoretical results on these properties under four representative survival models: DeepSurv, PC-Hazard, Nnet-Survival, and MTLR. For each model, we formalize its model structure and loss function, followed by two main theorems of the existence of interpolation and finite-norm interpolation.

All models below are expressed in a unified notation using \(z_\theta(x)\) for network outputs, 
where \(z_\theta(x)\) denotes either a scalar log-risk score or a vector of interval-specific logits 
\((z_{\theta,j}(x)\ \text{for interval } j )\). 
We use \(z_\theta(x)\) rather than \(f_\theta(x)\) to emphasize that, in survival settings, 
the loss is defined through {unobserved} risk scores or logits that enter the likelihood 
(e.g., Cox log-risk, discrete-time logits, PC-Hazard’s pre–softplus logit), 
whereas in classical supervised setups \(f_\theta(x)\) typically denotes a predictor 
for an {observed} target. Moreover, since all four models considered are likelihood-based, the loss functions are all negative log-likelihood.

\subsection{DeepSurv}
DeepSurv \citep{katzman2018deepsurv} extends the classical Cox proportional hazards model by representing the log-risk score $z_\theta(x_i)$ through a neural network.
The hazard function is
\[
\lambda(t\mid x_i) = \lambda_0(t)\exp(z_\theta(x_i)),
\]
where $\lambda_0(t)$ is the unspecified baseline hazard.
Because $\lambda_0(t)$ cancels out in the partial likelihood,
optimization depends only on the relative ordering of predicted risks.
The negative log partial likelihood is defined as
\[
\mathcal{L}_{\text{DeepSurv}}(\theta)= -\sum_{i:\delta_i=1}
\Big[
z_\theta(x_i)
- \log \sum_{j \in \mathcal{R}_i} \exp(z_\theta(x_j))
\Big],
\]
where $\mathcal{R}_i$ denotes the risk set at time $T_i$.
This loss function encourages higher risk scores for subjects who experience earlier events.

\begin{theorem}[Existence of interpolation in DeepSurv]
\label{thm:deepsurv-interp}
DeepSurv admits an interpolation. 
\end{theorem}
If the training data permit risk-set dominance, that is, there exists a sequence of scoring functions $\{\tilde{z}_{\theta}\}_{t>0}$, which is indexed by a scaling parameter (or model capicity) t such that for every event $i$,
\[
\min_{j\in\mathcal R_i\setminus\{i\}}
\big\{ \{\tilde{z}_{\theta}(x_i)\}_t - \{\tilde{z}_{\theta}(x_j)\}_t \big\} \;\xrightarrow[]{t\to\infty}\; +\infty,
\]
then $\inf_{\theta}\mathcal L_{\mathrm{DeepSurv}}(\theta)=0$.
The infimum is not attained at any finite parameter value unless every risk set is a singleton, 
but it is asymptotically achievable through increasingly dominant risk scores.

While interpolation is theoretically approachable by infinitely separating risk scores, achieving it with bounded parameters poses a distinct geometric challenge. 
The existence of finite-norm interpolation ultimately depends on how the last layer of the deep neural network translates the shared embedding into logits that determine the survival likelihood. We next introduce and analyze the geometry of the shared embedding.

For all neural survival architectures considered, the last layer of the network can be decomposed as
\[
\phi(x) = g(W f(x) + b),
\]
where $f(x)\in\mathbb{R}^u$ denotes the nonlinear feature representation
generated by preceding hidden layers, $W\in\mathbb{R}^{q\times u}$ and $b\in\mathbb{R}^{q}$
belong to the final linear layer, and $g(\cdot)$ (eg. softmax, sigmoid, softplus) is the activation function, such as softplus, sigmoid, or softmax. Here, $q=1$ for DeepSurv and $q=m$ for PC-Hazard, Nnet-Survival and N-MTLR, where $m$ is the number of time intervals. The representation $f(x)$ is shared across all time intervals, so this structure is referred to as a {shared embedding}. This shared-embedding structure is standard in survival neural networks, where the temporal component (hazard or cumulative score) is modeled through linear readouts of a common latent representation.

Because $W f(x)+b$ is linear in $W$, and because $g(\cdot)$ is strictly monotonic in each coordinate,
the margin that determines the training loss is fully governed by the linear logits $z_i=W f(x_i)+b$. Here we use the term “logits” generically to denote the pre-activation linear outputs 
$z_i$, which play the same geometric role across all models. In Nnet-Survival these correspond to log-odds, in PC-Hazard to log-relative risk, and in N-MTLR to per-interval scores.

The classic norm-margin tradeoff for linear classifiers \citep{bartlett2002rademacher,soudry2018implicit} thus can be extended to the last linear layer, since the margin governing the survival likelihood is determined solely by the logits 
$z_i$ (assuming differentiable and strictly monotonic activations). This extension holds at the level of the logit geometry, regardless of the specific likelihood form.
Intuitively, a large separation in predicted logits requires proportionally
large weight norms, as quantified in the following lemma.

\begin{lemma}[Margin budget under shared embedding for DeepSurv]\label{lemma:margin-budget}
Let the last layer adopt a shared embedding $f(x)$ be $\phi(x)=W f(x)+b$, with
$W\in\mathbb{R}^{1\times u}$, $f:\mathbb{R}^d\to\mathbb{R}^u$. 
Assume there exists $\gamma>0$ such that for every event subject $i$ and $j \in R_i \backslash \{i\}$
\begin{equation}\label{eqn:budget_deepsurv} 
\phi(x_i)-\phi(x_j)\ \ge\ \gamma.
\end{equation} Then $W$ must satisfy the following operator–norm lower bound
\begin{equation}\label{eqn:LB_deepsurv}
\|W\|_2\ \geq\ \frac{\gamma}{ \max_{i,j\in R_i\backslash \{i\} }\|f(x_i) - f(x_j)\|_2}. 
\end{equation}
\end{lemma}

Lemma~\ref{lemma:margin-budget} extends the classical result of
linear binary classification \citep{bartlett2002rademacher} to the shared-embedding setting used in neural survival models.
It quantifies how the required weight norm scales with the logit separation
needed to achieve a given margin $\gamma$. Building on the above lower bound of $W$, we next study the finite-norm interpolation for DeepSurv.

\begin{theorem}[Absence of finite-norm interpolation in DeepSurv]\label{thm:deepsurv-no-fg}
DeepSurv does not admit finite-norm interpolation.
\end{theorem}

DeepSurv’s partial likelihood depends only on relative log-risk differences within each risk set. Achieving near-zero loss thus requires exponentially growing logit gaps, which, through the shared-embedding margin constraint, force the weight norm to diverge. Hence, interpolation is attainable only in the theoretical limit but not under any finite norm. To be more specific, let $\mathcal{L}^\ast=\inf \mathcal{L}(\theta)$ be the optimal loss, then there exist constants $c_1,c_0>0$ such that: 
if a parameter $\theta_\varepsilon$ satisfies
\[
\mathcal L_{\mathrm{DeepSurv}}(\theta_\varepsilon)-\mathcal L^\ast \le \varepsilon,
\]
then the sample-level logit margin satisfies
\[
\min_{i:\,\delta_i=1}\ \min_{j\in\mathcal R_i\setminus\{i\}}\ 
\big\{\, z_{\theta_\varepsilon}(x_i) - z_{\theta_\varepsilon}(x_j)\,\big\}
\ \ge\ c_1\log(1/\varepsilon) - c_0.
\]
By Lemma~\ref{lemma:margin-budget} , $\|W\|_2\to\infty$ as $\varepsilon\downarrow 0$, which violates the finite-norm interpolation. 

\subsection{PC-hazard}
The PC-Hazard model \citep{kvamme2019time} specifies a full continuous-time likelihood through a piecewise-constant hazard. 
It assumes the hazard function is constant within each of $m$ predefined intervals $(\tau_{j-1}, \tau_j],~j=1,\cdots,m$, with corresponding log-intensity functions $z_{\theta,j}(x)$:
\[\lambda(t\mid x)
= \sum_{j=1}^{m}
\eta_j(x)\,\mathbf{1}\{\tau_{j-1} < t \le \tau_j\},\]
\[\eta_j(x)=\log(1+e^{z_{\theta,j}(x)}),
\]
For models defined with time intervals such as PC-Hazard and N-MTLR (see \Cref{sec:mtlr}),
we denote by \(j(i)\) the index such that the observed time
\(T_i \in (\tau_{j(i)-1}, \tau_{j(i)}]\).
Since $\eta_j(x)\in(0,\infty)$, the model ensures positive hazards while keeping gradients stable.
The loss function $\mathcal{L}_{\text{PC-Hazard}}(\theta)$ is defined as
\begin{align*}
\sum_{i=1}^n \left[
    -\delta_i \log \eta_{j(i)}(x_i)
    + \rho_i\,\eta_{j(i)}(x_i)+
 \sum_{k=1}^{j(i)-1} \eta_k(x_i)
\right],
\end{align*}
where
\(
\eta_j(x) = \log(1 + e^{z_{\theta,j}(x)})\) and 
\(\rho_i = \frac{T_i - \tau_{j(i)-1}}{\tau_{j(i)} - \tau_{j(i)-1}}\) denotes the normalized exposure length within the event interval for event observation $i$.

This formulation provides a fully specified continuous-time likelihood,
making interpolation mathematically well-defined through exact likelihood maximization.

\begin{theorem}[Existence of interpolation in PC-Hazard]
\label{thm:pchazard}
Let \( r_j = \#\{\, i : j(i) \geq j \,\} \) denote the number of individuals at risk in interval \( j \).
Then PC-Hazard admits an interpolation with $d^*\leq \sum_{j=1}^{m} r_j$.
\end{theorem}

\Cref{thm:pchazard} establishes a sufficient condition on number of parameters of PC-Hazard, i.e., $d=\sum_{j=1}^mr_j$, for interpolating all training samples in the sense of Definition~\ref{def:interpolation_math}. 

To study the finite-norm interpolation of PC-Hazard, we need a lemma analogous to Lemma \ref{lemma:margin-budget} tailored for models involving time intervals.

\begin{lemma}[Margin budget under shared embedding for models with intervals]\label{lemma:margin-budget-interval}
Let the last layer adopt a shared embedding $f(x)$ be $\phi(x)=W f(x)+b$, with
$W\in\mathbb{R}^{m \times u}$, $f:\mathbb{R}^d\to\mathbb{R}^u$.
Assume there exists $\gamma>0$ such that for every event subject $i$,
\begin{equation}\label{eqn:budget_new} 
\phi(x_i)_{j(i)}-\max_{k\neq j(i)}\phi(x_i)_k\ \ge\ \gamma
\end{equation}where $\phi(x_i)_k$ denotes the $k$-th entry of the $m$-dimensional output vector $\phi(x_i)$. Then $W$ must satisfy the following operator–norm lower bound
\begin{equation}\label{eqn:LB_new}
\|W\|_2\ \geq\ \frac{\gamma}{ \sqrt{2}\max_i\|f(x_i)\|_2}.
\end{equation}
\end{lemma}

\begin{theorem}[Absence of finite-norm interpolation in PC-Hazard]\label{thm:pch-fg} PC-Hazard does not admit finite-norm interpolation.
\end{theorem}
Similar to DeepSurv, if $\theta_\varepsilon$ satisfies
\[
\mathcal L_{\mathrm{PC-Hazard}}(\theta_\varepsilon)-\mathcal L^\ast \le \varepsilon.
\]
then  $z_{\theta,k}(x_i)\!\to\!-\infty$ as $\varepsilon\!\downarrow\!0$ for censored and pre-event intervals,
which implies $\|W\|_2 \!\to\! \infty$ under any shared-embedding parameterization by Lemma \ref{lemma:margin-budget-interval}. As a result, PC-Hazard does not admit finite-norm interpolation.

\subsection{Nnet-Survival}
Nnet-Survival \citep{gensheimer2019nnet} discretizes follow-up time into $m$ intervals $(\tau_{j-1}, \tau_j],~j=1,\cdots,m$. 
For subject $i$ and interval $j$, define the event indicator
\[
y_{ij} =
\begin{cases}
1, & \text{if subject $i$ experiences the event in } (\tau_{j-1}, \tau_j],\\
0, & \text{otherwise.}
\end{cases}
\]
The network outputs logits $z_{\theta,j}(x_i)$, with conditional hazard probability \[
h_{ij}(\theta)
= P\!\big(T_i \in (\tau_{j-1}, \tau_j] \mid T_i > \tau_{j-1},\, x_i;\, \theta\big),\]
\[
h_{ij}(\theta)
= \sigma(z_{\theta,j}(x_i))
= \frac{1}{1 + e^{-z_{\theta,j}(x_i)}}.
\]
The loss function of Nnet-Survival, $\mathcal{L}_{\text{Nnet-Survival}}(\theta)$, is
\begin{align*}
- \sum_{j=1}^{m} \sum_{i=1}^{r_j}
\left[
    y_{ij}\log h_{ij}(\theta)
    + (1 - y_{ij})\log\!\big(1 - h_{ij}(\theta)\big)
\right],
\end{align*}
where $r_j$ denotes the number of individuals at risk in interval $j$.
Each time interval contributes an independent conditional likelihood term.
Interpolation corresponds to perfect prediction of all observed event probabilities across intervals.

\begin{theorem}[Existence of interpolation in Nnet-Survival]
\label{thm:nnet-interp}
Nnet-Survival admits an interpolation with $d^*\leq mn$.
\end{theorem}

\begin{theorem}[Absence of finite-norm interpolation in Nnet-Survival]\label{thm:nnet-no-fg}
Nnet-Survival does not admit finite-norm interpolation.
\end{theorem}

The reason for the absence of finite-norm interpolation in Nnet-Survival mirrors that of DeepSurv: there exists constants $c_1,c_0>0$ such that if $\theta_\varepsilon$ satisfies
\[
\mathcal L_{\mathrm{Nnet-Survival}}(\theta_\varepsilon)-\mathcal L^\ast \le \varepsilon,
\]
then the margin of within-subject logit separation satisfies
\[
z_{\theta,j}(x_i) - \max_{k\neq j(i)}z_{\theta,k}(x_i)
\ \ge\ c_1 \log(1/\varepsilon) - c_0.
\]
That is, making all per-cell residuals $h_{ij}-y_{ij}$ of order $O(\varepsilon)$ requires logits to polarize, so that each positive cell dominates negative cells by a margin of order $\log(1/\varepsilon)$. Then by Lemma~\ref{lemma:margin-budget-interval}, the required weight norm diverges, i.e. $\|W\|_2 \to \infty$ as $\varepsilon\downarrow 0$, which violates the finite-norm interpolation.

\subsection{N-MTLR}\label{sec:mtlr}

The N-MTLR model \citep{Fotso2018DeepNN} is also defined over discretized time intervals $(\tau_{j-1}, \tau_j],~j=1,\cdots,m$. 
The model introduces cumulative logits $\{z_{\theta,k}(x)\}_{k=1}^{m}$. The event probability for interval $j$ and the survival probability after $\tau_j$ are
\[
p_{\theta,j}(x)
= \frac{\exp(\sum_{k=j}^{m} z_{\theta,k}(x))}
{\sum_{l=1}^{m} \exp(\sum_{k=l}^{m} z_{\theta,k}(x))},\]
\[S_{\theta,j}(x) = \sum_{l=j+1}^{m} p_l(x),
\]
where $0 < p_{\theta,j}(x) < 1$ and $\sum_{j=1}^{m} p_{\theta,j}(x) = 1$.
The loss function of N-MTLR, $\mathcal{L}_{\text{N-MTLR}}(\theta)$, is 
\[
 -\sum_{i=1}^{n}
\left[
\delta_i \log p_{\theta,j(i)}(x_i)
+ (1 - \delta_i) \log S_{\theta,j(i)}(x_i)
\right].
\]
The cumulative structure of $\sum_{k=j}^{m} z_{\theta,k}(x)$ enforces monotonic survival probabilities
but also couples interval-specific logits,
restricting the possibility of perfect interpolation.

\begin{theorem}[Existence of interpolation in N-MTLR]
\label{thm:mtlr-interp}
N-MTLR admits an interpolation with $d^*\leq mn$. 
\end{theorem}

\begin{theorem}[Absence of finite-norm interpolation in N-MTLR]\label{thm:mtlr-no-fg}
N-MTLR does not admit a finite-norm interpolation. 
\end{theorem}
For the N-MTLR model with softmax-normalized cumulative scores,
achieving zero loss requires a logit separation
$\gamma(\varepsilon)\simeq \log(1/\varepsilon)$ across $m$ cumulative intervals.
Because the cumulative transformation scales the effective margin 
$\asymp\sqrt{m}$, the required weight norm diverges as
\[
\|W\|_2
\gtrsim
\frac{\log(1/\varepsilon)}{\sqrt{m}\ \max_i\|f(x_i)\|_2}
\ \xrightarrow[\varepsilon\to0]{}\ \infty,
\]
which violates the finite-norm interpolation.

\subsection{Summary}

\Cref{thm:deepsurv-interp,thm:pchazard,thm:nnet-interp,thm:mtlr-interp} confirm the existence of an interpolation of all four models. \Cref{thm:deepsurv-no-fg,thm:pch-fg,thm:nnet-no-fg,thm:mtlr-no-fg}, on the other hand, show the absence of finite-norm interpolation of all four models.  Moreover, Lemma \ref{lemma:margin-budget}, \ref{lemma:margin-budget-interval} reframe the finite-norm interpolation problem of survival models into a geometric margin problem of a linear classifier. The existence (or absence) of finite-norm interpolation follows from how each model’s activation function dictates the required logit margin, thereby linking loss minimization to the classical margin–norm tradeoff \citep{bartlett2002rademacher}.

\section{Simulation}\label{sec:simulation}

In this section, we perform simulation studies to empirically support the theoretical results in \Cref{sec:theory}. For all four models, we vary the number of parameters $d$ (i.e., model capacity) and examine the trajectories of both training and testing loss. This allows us to identify the interpolation points (where training loss plateaus) and to assess whether a second descent in test loss appears beyond this point. 

For DeepSurv, because its network output corresponds directly to the log-hazard function (or risk score) $z_\theta$ in the data-generating mechanism, we additionally evaluate whether the learned $z_\theta$ recovers the true log-hazard. While for PC-Hazard, Nnet-Survival, and N-MTLR, the network outputs do not parameterize the log-hazard in a comparable form, so such evaluation is skipped.

\subsection{Data Generating Mechanism}

We generate data under a Weibull baseline hazard with shape parameter \(\gamma = 0.7\) and uniform censoring \(C \sim \text{Unif}(0,0.8)\).
For PC-Hazard, Nnet-Survival, and N-MTLR, each simulated dataset contains \(n = 3{,}500\) individuals and \(p = 200\) covariates.
For DeepSurv, we use \(p = 60\) covariates to ensure stable convergence of the partial-likelihood optimization. 

Covariates follow an AR(1) correlation structure with parameter \(\rho = 0.6\):
\[
X \sim \mathcal{N}_p(0, \Sigma_\rho), \qquad (\Sigma_\rho)_{kl} = \rho^{|k-l|}.
\]
Among these, 50 variables are truly relevant with coefficients
\(\beta_k \sim \text{Unif}(-0.5, 0.5)\).
The true log-hazard function is
\[
\eta(x) = 0.31 \sum_{k \in \mathcal{S}} \beta_k \mathbf{1}(x_k > 0),
\]
where \(\mathcal{S}\) denotes the set of relevant features. The constant 0.31 serves as a scaling factor controlling the overall signal-to-noise ratio.
It was chosen to yield a moderate effect size such that the average event rate remains within a realistic range (around 55\% censoring).
This value is fixed and not tuned for performance.
Event times follow
\[
T = (-\log U / \exp(\eta(x)))^{1/\gamma}, \qquad U \sim \text{Unif}(0,1),
\]
and observed times are
\[
Y = \min(T, C, \tau), \qquad \delta = \mathbf{1}(T \le C, T \le \tau),
\]
with \(\tau = 0.6\).
This setup defines a  piecewise-linear risk surface, allowing precise control of the signal-to-noise ratio and the true function norm \(\|\eta(x)\|_2  \), which empirically equals 0.346 in our test dataset and 0.356 in our training dataset.
\begin{remark}
For PC-Hazard, Nnet-Survival, and N-MTLR we use \(p=200\) covariates, whereas for DeepSurv we set \(p=60\).
This design is intentional and aims at visualizing the entire capacity–loss trajectory (from the under-parameterized regime to the interpolation threshold and further into the over-parameterized regime) within a moderate range of hidden-layer widths.
Because DeepSurv produces a scalar risk score, the effective model capacity needed to reach interpolation scales with both the input dimension and the hidden width. Using \(p=60\) shifts the interpolation point to smaller widths, making the double-descent pattern observable without resorting to extreme network sizes.
This choice changes only the horizontal scaling of the capacity axis; it does not alter the qualitative phenomena (existence of interpolation, secondary descent or divergence) that our analysis focuses on.
For the other models, the multi-output heads (interval-wise hazards or cumulative probabilities) already induce sufficient effective capacity at \(p=200\), so the key behaviors emerge without reducing \(p\).  
\end{remark}

\subsection{Evaluation Metrics}

To examine the empirical manifestation of double-descent and its alignment with theoretical predictions,
we monitor the following quantities across increasing model capacities:
\begin{enumerate}
\item \textit{Training loss:} recorded to verify the existence of interpolation points and optimization completeness: its monotonic decrease ensures that observed non-monotonicity in test loss is not due to incomplete training;
    \item \textit{Test loss:} the primary indicator of double-descent behavior, reflecting generalization loss as capacity grows beyond the interpolation point;
    \item \textit{Norm comparison:} for DeepSurv, the \(L_2\) norm of the estimated \(z_\theta(x)\) is compared against the true function norm \(\|\eta(x)\|_2\), providing insights into the behavior of the deep neural network in DeepSurv and how it affects double-descent.
\end{enumerate}

Theoretical results imply that as capacity increases, the model space expands in a nested manner and the training loss should not rise if optimization is complete. 
Hence, any non-monotonic pattern observed in the test loss reflects genuine post-interpolation generalization effects rather than underfitting.
All models are trained without explicit regularization (e.g., dropout, weight decay, or penalty terms),
so that any observed loss pattern arises purely from the model’s own loss formulation rather than external regularization effects.

\subsection{DeepSurv}

\begin{figure*}[t]
\vspace{-0.1in}
\centering
    \centering
    \includegraphics[width=0.99\linewidth]{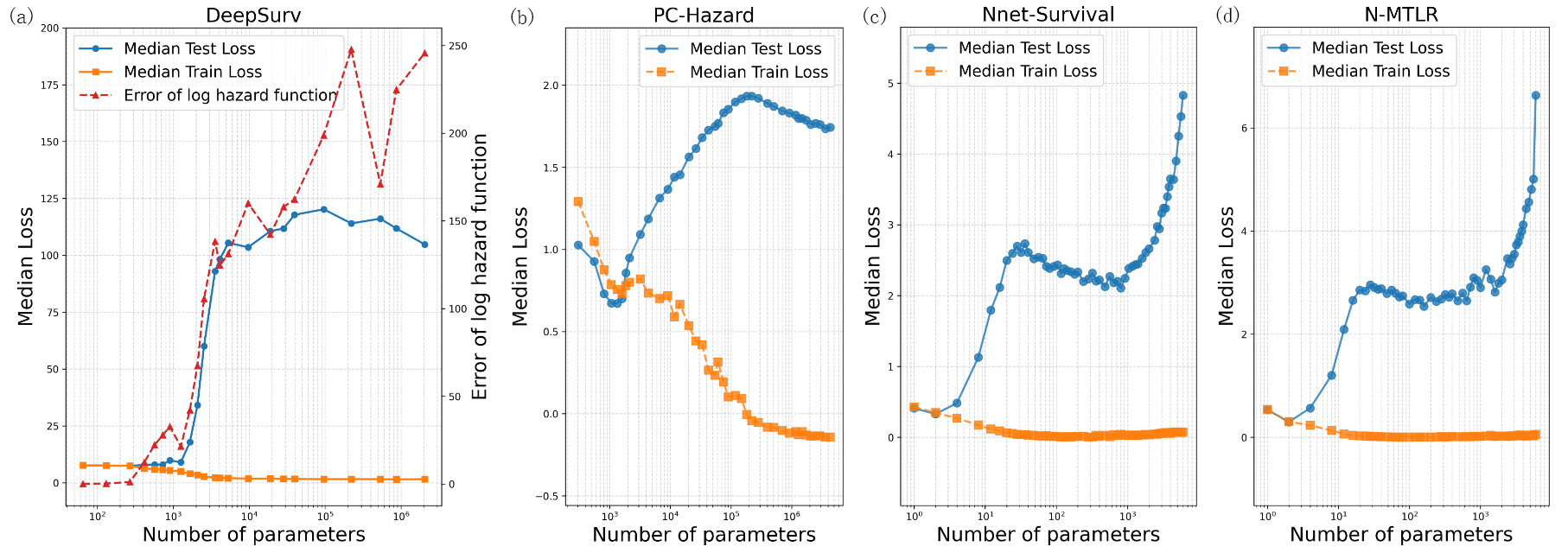}
\caption{Training and test losses of DeepSurv (a), PC-Hazard (b), Nnet-Survival (c), and N-MTLR (d) versus the number of parameters in the models. For DeepSurv, the error of the estimated log-hazard function $z_\theta$ is also provided.
}
\vspace{-0.15in}
\label{fig:results}
\end{figure*}

\Cref{fig:results}(a) shows that as model capacity increases, the test loss decreases initially, reaches a local minimum, 
then rises sharply near the interpolation threshold, followed by a mild downward drift 
that stabilizes at a substantially higher level than the first minimum. 

This brief “quasi-descent” does not reflect true finite-norm stability, which is consistent with our theoretical result that DeepSurv admits interpolation (\Cref{thm:deepsurv-interp}) but not finite-norm interpolation (\Cref{thm:deepsurv-no-fg}). 
The apparent plateau corresponds to the optimizer entering a region of nearly flat gradients, 
where the partial likelihood continues to approach its infimum but only through exponentially increasing logit separations $ z_{\theta}(x_i) - z_{\theta}(x_j)$  and diverging weight norms. Interestingly, clear double-descent behavior has often been observed in models that admit finite-norm interpolation, whereas DeepSurv represents a nuanced edge case.

Another interesting observation is that in the training dataset, the true negative partial log-likelihood (NPLL) computed from the data-generating mechanism is approximately 6.646, while the empirical NPLL of the training dataset approaches 0 (infimum of the loss function) asymptotically without converging under bounded parameter norms. 
During this process, the estimated \(L_2\) norm of the network output \(z_\theta(x)\) remains consistently higher than the true function norm and increases rapidly once the empirical NPLL crosses the true NPLL as shown by the red dash line in Figure~\ref{fig:results}(a) (representing the deviation between the \(L_2\) norm of \(z_\theta(x)\)and that of the true log-hazard function). This discrepancy highlights that the optimization objective (NPLL) is not aligned with the data-generating truth: driving the loss toward zero no longer improves fidelity to the underlying event process but merely inflates relative risk margins, rendering the tail region of optimization practically meaningless.

Notably, the test loss does not continue to increase indefinitely and enter an ``high plateau''.
This apparent plateau does not represent a meaningful interpolating regime but rather an optimization stall induced by gradient saturation. 
The resulting curve only superficially resembles double descend, but its origin lies in the asymptotic flattening of the partial-likelihood surface rather than any restoration of generalization or finite-norm interpolation.

\subsection{PC-Hazard}
PC-Hazard exhibits a pronounced three-phase double-descent trajectory (Figure~\ref{fig:results}(b)). 
As network capacity increases, the training loss decays monotonically and stabilizes at the interpolation point as confirmed in \Cref{thm:pchazard}. In contrast, the test loss first decreases, indicating improved data fit, 
then rises from the sweet spot until the interpolation threshold, and finally declines again to a stable plateau at larger capacities. This “decrease–increase–decrease” sequence represents a canonical double-descent pattern.

Unlike DeepSurv, PC-Hazard is trained under a fully specified continuous-time likelihood, 
where each event interval contributes an explicit likelihood term through the softplus-transformed log-rate. 
As the model approaches interpolation, censored intervals uniformly drive their corresponding logits toward negative infinity, 
causing the loss to approach its infimum exponentially fast with respect to the logit norm. 
This rapid attenuation produces a broad near-flat region in the loss landscape, 
allowing the network to reach an effectively near-interpolating regime within finite norms even though strict finite-norm interpolation is unattainable (\Cref{thm:pch-fg}).

Beyond this region, further increases in model capacity do not destabilize the optimization trajectory. 
Gradients remain well-behaved, and both training and test losses converge to stable levels, suggesting that the PC-Hazard
provides an implicit regularization effect through its piecewise hazard parameterization. 
The observed second descent therefore arises not from bounded-norm interpolation but from the implicit regularization and exponential saturation of the PC-Hazard loss, 
which enables the model to approximate its loss infimum in practice.

\subsection{Nnet-Survival}

Nnet-Survival displays a divergent rather than stabilizing pattern (Figure~\ref{fig:results}(c)). 
As model capacity increases, the training loss decreases monotonically toward zero at the interpolation threshold as confirmed by \Cref{thm:nnet-interp}.
In contrast, the test loss behaves very differently: it first declines, reaches a shallow plateau, then rises sharply to a higher plateau, and finally explodes once the network becomes highly over-parameterized. No secondary descent is observed on the test curve, suggesting that the model fails to regain generalization after interpolation.

This explosive rise reflects the geometric limitation of the discrete-time logistic formulation. 
Each interval-specific likelihood term is bounded by the sigmoid transformation, which restricts the attainable probability mass within each interval. To further reduce the loss, the network must therefore push its logits toward extreme magnitudes, leading to rapid margin inflation and curvature blow-up in the loss landscape. As a result, the optimization dynamics remains unstable in the over-parameterized regime, which prevents the emergence of a complete double-descent shape.

\subsection{N-MTLR}


N-MTLR exhibits a similar overall trajectory to Nnet-Survival (Figure~\ref{fig:results}(d)). 
As model capacity increases, the training loss decreases steadily and hit the interpolation threshold confirmed by \Cref{thm:mtlr-interp}. In contrast, the test loss initially falls, levels off over a moderate range, 
and then rises again once the network becomes heavily over-parameterized. 
No secondary descent is observed, suggesting that the model fails to recover generalization after interpolation.

This behavior reflects the structural constraint imposed by the cumulative-logit parameterization. 
Because all interval-specific logits are coupled through the monotonic survival constraint, 
the model cannot freely adjust interval risks once the shared embedding saturates. 
As additional neurons are introduced, this coupling causes gradients across intervals to point in partially opposing directions, preventing uniform gradient decay and leading to a slow but persistent decline of the training loss. 
Consequently, the test loss increases even as the training loss continues to fall, producing a smooth but divergent trajectory without a coherent second descent. 
Ultimately, N-MTLR shares the same qualitative limitation as Nnet-Survival: its discrete-time formulation induces coupled gradients that hinder stable near-interpolation behavior and preclude observable double-descent.

\subsection{Discussion}

Theoretically, none of the four survival neural networks admits finite-norm interpolation: 
in all cases, achieving the loss infimum requires logits to diverge to $\pm\infty$. 
Nevertheless, their generalization curves differ markedly, reflecting how distinct loss geometries 
govern near-interpolation behavior and secondary descents. 
Although the following hierarchy is not formal in the mathematical sense, it provides a conceptual framework linking the existence and geometry of interpolation to the observability of double-descent in survival models:

\begin{itemize}
\item[(i)] \textit{Existence of interpolation.} 
A theoretical existence of interpolation threshold supports the potential turning point of the test loss that may lead to a double-descent.

\item[(ii)] \textit{Existence of finite-norm interpolation.}  
If the loss minimum were attainable within bounded norms, 
it would yield a locally smooth, low-curvature region where implicit regularization could operate, 
producing a clear and stable second descent.

\item[(iii)] \textit{No finite-norm interpolation with fast-decaying near-interpolation.} 
If the loss decays exponentially fast toward its infimum along a coherent gradient direction 
(e.g., PC-Hazard), a broad near-flat region emerges in the loss landscape. 
Optimization stabilizes within finite norms, producing a clear second descent in test loss even without a bounded interpolating solution.

\item[(iv)] \textit{No finite-norm interpolation with heterogeneous near-interpolation.} 
When gradients decay unevenly or in opposing directions across coupled logits (e.g. DeepSurv, Nnet-Survival, N-MTLR), 
no globally flat region forms. 
Scale inflation replaces curvature flattening, yielding at most a weak or transient test-loss recovery.

\item[(v)] \textit{No finite-norm interpolation with pre-interpolation instability.} 
If coupling among logits prevents consistent margin growth (e.g. Nnet-Survival, N-MTLR), the model never reaches a stable near-interpolation regime. 
Optimization noise dominates, and both training and test losses degrade monotonically.
\end{itemize}

Together, these regimes show that observable double-descent behavior in survival models does not only depend on the existence of interpolation mathematically or finite-norm interpolation itself, 
but also on how the loss geometry and loss gradient coupling shape the approach toward its asymptotic infimum.

\noindent\textit{Empirical patterns across models.} Even when a second descent appears, as in PC-Hazard or to a lesser extent DeepSurv, it does not guarantee improved generalization compared with the sweet spot in the traditional framework. Increasing model capacity beyond the near-interpolation region provides little benefit as past-interpolation test loss is always higher than the one on the sweet spot. Practitioners should be cautious for excessive over-parameterization in survival neural networks,
as the resulting complexity may amplify estimation noise without yielding meaningful performance gains.

\noindent\textit{Dependence on the loss.} 
The presence or absence of double-descent-like behavior in survival neural networks is fundamentally loss-specific. 
Each likelihood formulation imposes distinct curvature, saturation, and coupling properties that shape how optimization responds to increasing model capacity. 
Losses derived from smooth, monotone link functions tend to produce coherent gradient flows and relatively stable near-interpolation behavior. 
In contrast, losses built from bounded transformations such as sigmoids or cumulative logits create competing gradient directions across time intervals, 
preventing stable convergence once the network becomes highly over-parameterized. 
Thus, the formulation and the geometry of the loss surface govern whether stable near-interpolation behavior and secondary descents can emerge in practice.

\noindent\textit{Unobservable optimization targets.} A key implication of these findings is that the optimization targets in survival neural networks are inherently unobservable from data. Each loss function operates on latent quantities that cannot be directly measured, such as log-risk scores or interval-specific logits. 
DeepSurv provides a particularly clear example: its network estimates a log-hazard ratio, yet the true log-hazard function is not identifiable from censored data. 
As shown in our simulations, the model continues to reduce training loss even after surpassing the empirical true loss computed from the data-generating mechanism. 

From a broader perspective, these findings suggest that the double-descent phenomenon in survival neural networks is not merely a numerical artifact of over-parameterization.
It reflects a more fundamental limitation: when the target of learning is itself unobservable, optimization can proceed indefinitely without converging toward the underlying truth.
In this sense, survival neural networks “optimize for what cannot be seen”, and this intrinsic misalignment defines their generalization boundaries.

\section{Future work}
This study theoretically established the existence of interpolation and the absence of finite-norm interpolation in all four representative survival models, showing that their distinct double-descent behaviors stem from how each loss geometry approaches its asymptotic infimum.

Building upon our work, we anticipate several important future directions. First, while our theory establishes the existence or absence of interpolation and finite-norm interpolation, it does not guarantee the existence or absence of double-descent. A theoretical characterization of when double-descent is guaranteed remains open even for standard regression problem. Second, even when a double-descent exists, whether the second local minimum is higher or lower than the first one, i.e., the classical sweet spot, remains unresolved for both survival and regression models. This comparison is essential in determining whether over-parametrization is beneficial or not, a key practical question for deep survival models. The third future direction focuses on coupling, to examine how interactions among logits determine the efficiency with which models approximate the loss infimum under bounded norms. Finally, analyzing how coherent or conflicting of gradient directions near the interpolation region may yield practical diagnostics for model selection, informing when increasing model capacity will improve or degrade optimization stability and generalization.

\appendix
\section{Proofs}
\label{appendix_A}


In this appendix we prove the following theorem from
Section~\ref{sec:theory}:

\subsection{ Proof of Theorem~\ref{thm:deepsurv-interp}~(Existence of interpolation in DeepSurv)}
\label{app:deepsurv-interp}

\begin{proof}
Let
\[
\mathcal L_{\mathrm{DeepSurv}}(\theta)
=\sum_{i:\delta_i=1}\ell_i(\theta),~
\ell_i(\theta)
=-\Big[z_\theta(x_i)-\log\!\sum_{j\in\mathcal R_i}e^{z_\theta(x_j)}\Big]
=\log\!\Big(1+\sum_{j\in\mathcal R_i\setminus\{i\}} e^{-[z_\theta(x_i)-z_\theta(x_j)]}\Big),
\]
and define the minimal margin within each risk set as
\[
m_i(z_\theta)=\min_{j\in\mathcal R_i\setminus\{i\}}[z_\theta(x_i)-z_\theta(x_j)],\quad
m_i(z_\theta)=+\infty\ \text{if}\ \mathcal R_i=\{i\}.
\]
Let $\mathcal P=\{(i,j):\ \delta_i=1,\ j\in\mathcal R_i\setminus\{i\}\}$ be the set of comparable pairs. \(z_\theta(x)\) denotes the scalar log-risk score, which is the output from the neural net. We split the proof into three steps.

\noindent \textbf{Step 1: Approximation of a separating risk function.}
Assume there exists a continuous log-risk function \(f:\mathbb R^p\to\mathbb R\)
such that \(f(x_i)>f(x_j)\) for all \((i,j)\in\mathcal P\). Here $f$ represents the underlying continuous log-risk function assumed in nonlinear extensions of the Cox model \citep{katzman2018deepsurv}.
Such models impose no structural form on $f$ other than continuity.
Define the minimal pairwise gap
\[
\Delta_{\min}=\min_{(i,j)\in\mathcal P}\{f(x_i)-f(x_j)\}>0.
\]

By the Universal Approximation Theorem~\citep{cybenko1989approximation,hornik1991approximation}, for any continuous function \( f \) on the a compact set $K\subset\mathbb{R}^p$, for any $\varepsilon>0$, there exists a neural network
\(z_\theta:\mathbb R^p\to\mathbb R\) where $\theta\in\mathbb{R}^{d(\varepsilon)}$ for some finite $d(\varepsilon)$ such that
\(|z_\theta(x)-f(x)|<\varepsilon\) for all $x\in K$.
Hence, fix $\varepsilon < \frac{\Delta_{\min}}{2}$, then there exists a neural network $z_\theta$ where $\theta\in\mathbb{R}^{d(\varepsilon)}$  such that for every \((i,j)\in\mathcal P\),
\[
z_\theta(x_i)-z_\theta(x_j)
\ge (f(x_i)-f(x_j))-2\varepsilon
\ge \Delta_{\min}-2\varepsilon>0,
\]
which establishes risk-set separability:
each event has strictly higher score than all members of its risk set. \\
\textbf{Step 2: Construction of a dominant sequence.}
Let \(z_\theta\) be the risk-set–separating score function constructed in step 1,
and define its global minimal margin
\[
\gamma=\min_{(i,j)\in\mathcal P}\{z_\theta(x_i)-z_\theta(x_j)\}>0.
\]
Consider the scaling path \(z_\theta(\cdot)^{(t)}=t\,z_\theta(\cdot)\), \(t>0\).
Then for every event \(i\),
\[
m_i\big(z_\theta^{(t)}\big)
=\min_{j\in\mathcal R_i\setminus\{i\}}\big(z_\theta(x_i)^{(t)}-z_\theta(x_j)^{(t)}\big)
\ge t\,\gamma \to +\infty
\quad(t\to\infty),
\]
so \(\{z^{(t)}_\theta\}\) forms a dominant sequence
whose margins diverge simultaneously for all events.\\
\textbf{Step 3: Dominance implies vanishing loss.}
Using \(e^{-[z_\theta(x_i)-z_\theta(x_j)]}\le e^{-m_i(z_\theta)}\) and letting \(k_i=|\mathcal R_i|-1\),
\[
\ell_i(\theta)\le \log\!\Big(1+k_i\,e^{-m_i(z_\theta)}\Big).
\]
Let $\theta^{(t)}$ denote any parameter vector whose induced logits satisfy
$z_{\theta}(x)^{(t)} = t\, z_\theta(x)$.
For \(z^{(t)}_\theta\) as above,
\[
\ell_i\big(\theta^{(t)})
\le \log\!\Big(1+k_i\,e^{-t\gamma}\Big)
\;\xrightarrow[]{t\to\infty}\; 0,\quad
\mathcal L_{\mathrm{DeepSurv}}\big(\theta^{(t)})
=\sum_{i:\delta_i=1}\ell_i\big(\theta^{(t)}) \;\xrightarrow[]{t\to\infty}\; 0.
\]
Therefore \(\inf_{\theta\in\mathbb{R}^{d(\varepsilon)}}\mathcal L_{\mathrm{DeepSurv}}(\theta)=0.\) and hence $d^*\leq d(\epsilon)<\infty$ and DeepSurv admits an interpolation.
\end{proof}

\subsection{Proof of Lemma~\ref{lemma:margin-budget}~(Margin budget under shared embedding for DeepSurv)}
\label{app:margin-budget}

\begin{proof}
Let the last-layer output be $\phi(x)=W f(x)+b$ with $W\in\mathbb{R}^{1\times u}$.
For any comparable pair $(i,j)\in\mathcal P=\{(i,j):\delta_i=1,\ j\in R_i\setminus\{i\}\}$,
the margin condition implies
\[
\phi(x_i)-\phi(x_j)
=(W f(x_i)+b)-(W f(x_j)+b)
= W\big(f(x_i)-f(x_j)\big)
\;\ge\; \gamma.
\]
Define $a_{ij}=f(x_i)-f(x_j)$, then the above inequality becomes $W a_{ij} \;\ge\; \gamma.$

By the operator-norm inequality,
$\|W\|_2\,\|a_{ij}\|_2\geq \|W a_{ij}\| \geq \gamma$, that is, $\|W\|_2\geq \frac{\gamma}{\|a_{ij}\|}$.

As a result,
\[
\|W\|_2\geq \min_{(i,j)\in\mathcal{P}}\frac{\gamma}{\|a_{ij}\|}=\frac{\gamma}{\max_{(i,j)\in\mathcal{P}}\|a_{ij}\|}=\frac{\gamma}{\max_{(i,j)\in\mathcal{P}}\|f(x_i)-f(x_j)\|}.
\]
which is the desired margin-norm lower bound. 
\end{proof}

\subsection{Proof of Theorem~\ref{thm:deepsurv-no-fg}~(Absence of finite-norm interpolation in DeepSurv)}
\label{app:deepsurv-no-fg}

\begin{proof}

By Definition~\ref{def:finite-interpolation}, finite-norm interpolation requires that the loss infimum $L^*$ be attainable within a bounded parameter set. Thus it suffices to show that any sequence $\theta_\varepsilon$ satisfying
\[
\mathcal L_{\mathrm{DeepSurv}}(\theta_\varepsilon) - L^* \le \varepsilon
\]
must have diverging norm as $\varepsilon \to 0$. We first show that near-optimal loss implies a margin of order $\log(1/\varepsilon)$, and then apply Lemma~\ref{lemma:margin-budget} to translate this margin growth into divergence of the last-layer weight norm, which is a component of the full parameter norm. Consequently, the parameter norm must diverge as $\varepsilon \to 0$, implying that the loss infimum cannot be attained within any bounded parameter set.

\noindent\textbf{Step 1: Per-event loss bound.}
Define
\[
s_i(\theta):=\sum_{j\in\mathcal R_i\setminus\{i\}} e^{z_\theta(x_j)-z_\theta(x_i)} \;\ge\;0,
\]
so that $\ell_i(\theta)=\log(1+s_i(\theta))$.
Then if
\[
\mathcal L_{\mathrm{DeepSurv}}(\theta_\varepsilon)-\mathcal L^\ast
=\sum_{i:\delta_i=1}\ell_i(\theta_\varepsilon)\le\varepsilon,
\]
 we must have each individual term $0\leq \ell_i(\theta_\varepsilon)\le\varepsilon$ for every event $i$. 
Thus
\[
s_i(\theta_\varepsilon)
= e^{\ell_i(\theta_\varepsilon)}-1
\le e^{\varepsilon}-1.
\]
\noindent\textbf{Step 2: Small loss forces large logit margin.}
For any $j\in\mathcal R_i\setminus\{i\}$,
\[
e^{z_\theta(x_j)-z_\theta(x_i)}
\le s_i(\theta_\varepsilon)
\le e^{\varepsilon}-1,
\]
hence
\[
z_\theta(x_i)-z_\theta(x_j)
\;\ge\;
\log\!\frac{1}{e^{\varepsilon}-1}.
\]
For $0<\varepsilon\le\log 2$, we have $e^{\varepsilon}-1\le 2\varepsilon$ and therefore
\[
z_\theta(x_i)-z_\theta(x_j)
\;\ge\;
\log\!\frac{1}{2\varepsilon}
=\log(1/\varepsilon)-\log 2.
\]
Taking the minimum over all events $i$ and $j\in\mathcal R_i\setminus\{i\}$ gives
\[
\min_{i:\delta_i=1}\ \min_{j\in\mathcal R_i\setminus\{i\}}
\big\{z_\theta(x_i)-z_\theta(x_j)\big\}
\;\ge\;
\log(1/\varepsilon)-\log 2
=: \gamma(\varepsilon).
\]
\noindent\textbf{Step 3: Margin forces norm blow-up under shared embedding.}
Express the last layer as $z_\theta(x)=W f(x)+b$ with shared embedding $f(\cdot)$.
By Lemma~\ref{lemma:margin-budget}, any $W$ realizing a uniform margin
$\gamma(\varepsilon)$ across all comparable pairs satisfies
\[
\|W\|_2 \;\geq\;
\frac{\gamma(\varepsilon)}{\displaystyle
\max_{(i,j)\in\mathcal P}\|f(x_i)-f(x_j)\|_2}\xrightarrow[]{\varepsilon\to0}\infty.
\]
 
\end{proof}

\subsection{Proof of Theorem~\ref{thm:pchazard}~(Existence of interpolation in PC-Hazard)}
\label{app:pchazard-interp}

\begin{proof}
We consider the per-sample loss for PC-Hazard,
\[
\ell_i(\theta) = -\delta_i \log {\eta}_{j(i)}(x_i) 
+ \rho_i {\eta}_{j(i)}(x_i)
+ \sum_{k=1}^{j(i)-1} {\eta}_k(x_i),
\]
where ${\eta}_k(x_i)=\log(1+e^{z_{\theta,k}(x_i)})$ denotes the
softplus-transformed log-rate. $z_{\theta,k}(x_i)\ $ denotes the neural net output for interval $k$ and subject $i$. 
We treat ${\eta}_k(x_i)\in(0,\infty)$ as free variables and minimize $\ell_i(\theta)$
without yet imposing a parametric form on $\theta$. In the remainder of the proof, we construct a finite-dimensional solution by working directly with free variables $\eta_k(x_i)$ and logits $z_{\theta,k}(x_i)$ in three steps.\\
\textbf{Step 1. Optimal values of ${\eta}_{j(i)}(x_i)$.} For intervals $k<j(i)$, choose ${\eta}_k(x_i)\le \varepsilon$, yielding 
$\sum_{k<j(i)}{\eta}_k(x_i)\le (j(i)-1)\varepsilon \;\xrightarrow[]{\varepsilon\to 0}\; 0$.
For the event interval $k=j(i)$, define $f(\eta)=-\delta_i\log \eta+\rho_i\eta$, which is strictly convex for $\eta\in(0,\infty)$ with unique minimizer $\eta^*=\delta_i/\rho_i$.
Hence:
\begin{itemize}
\item If $\delta_i=0$, choose $\eta_{j(i)}^*(x_i)=\varepsilon$ so that $f(\eta_{j(i)}^*(x_i))=\rho_i\varepsilon\to0$.
\item If $\delta_i=1$, the optimum is $\eta_{j(i)}^*(x_i)=\delta_i/\rho_i$ and $f(\eta_{j(i)}^*(x_i))=1+\log\rho_i<\infty$.
\end{itemize}
Thus censored contributions can be made arbitrarily small, while event contributions attain their finite per-sample minima.\\
\textbf{Step 2. Realizing ${\eta}_{k}(x_i)$ via $z_{\theta,k}(x_i)$.} Since ${\eta}_k(x_i)=\log(1+e^{z_{\theta,k}(x_i)})$ is strictly increasing in $z_{\theta,k}(x_i)$,
for any desired $\eta^*\in(0,\infty)$ there exists a unique $z^*$ satisfying ${\eta}(z^*)=\eta^*$. 
Hence, we choose ${\eta}_k(x_i) = \varepsilon$, which implys $z_{\theta,k}(x_i)\;\xrightarrow[]{\varepsilon\to 0}\; -\infty$ for $k< j(i)$. For $k= j(i)$:
\begin{itemize}
\item For $\delta_i=0$, $\eta_{j(i)}^*(x_i)=\varepsilon$ gives $z_{\theta,j(i)}^*(x_i)=\log(e^{\varepsilon}-1)\to -\infty$.
\item For $\delta_i=1$, $\eta_{j(i)}^*(x_i)=\delta_i/\rho_i$ gives $z_{\theta,j(i)}^*(x_i)=\log(e^{1/\rho_i}-1)<\infty$.
\end{itemize}
Thus, for each $(i,k)$ pair with $k\le j(i)$, we can specify $z_{\theta,k}(x_i)$ that is finite for any fixed $\varepsilon>0$, taking $\varepsilon \to 0 $ drives censored and pre-event terms to their infimum. 

\noindent\textbf{Step 3. Finite-dimensional interpolation.}
To achieve the infimum of the loss, Steps~1-2 require specifying logit 
$z_{\theta,k}(x_i)$ for each pair $(i,k)$ with $k \le j(i)$. 
For each interval $k$, let $r_k = \#\{\, i : j(i)\ge k \,\}$,
the number of individuals at risk in interval $k$.  
Hence the total number of required logit values is
$d=\sum_{k=1}^m r_k < \infty$. 

Since $z_{\theta,k}(x_i)$ is a map indexed by $\theta$, matching a
prescribed set of $d=\sum_{k=1}^m r_k$ logit values requires at most $d$
degrees of freedom.  In other words, we can always choose $\theta\in\mathbb{R}^d$ so that each coordinate of $\theta$ directly specifies one required logit value $z_{\theta,k}(x_i)$ and all $d$ constraints are satisfied simultaneously.
Therefore the $d$-dimensional parameter space is sufficient to realize all logit assignments  so that $\inf_{\theta\in\mathbb{R}^{d}}\mathcal{L}(\theta)=0$.
Therefore PC-Hazard admits an interpolation with
\[
d^* \le d = \sum_{k=1}^m r_k .
\]
\end{proof}

\subsection{Proof of Lemma~\ref{lemma:margin-budget-interval}~(Margin budget under shared embedding for models with intervals)}
\label{app:margin-budget-interval}

\begin{proof}
Let the last-layer logits be $\phi(x)=Wf(x)+b$, where $W\in\mathbb{R}^{m\times u}$ and
$f:\mathbb{R}^d\to\mathbb{R}^u$.  
Absorb the bias term by defining the augmented embedding 
$\tilde f(x)=(f(x)^\top,1)^\top$ and 
$\tilde W=(W,\ b)$, so that $\phi(x)=\tilde W \tilde f(x)$.  
Renaming $(\tilde W,\tilde f)$ back to $(W,f)$, we may assume $\phi(x)=Wf(x)$.
For an event subject $i$, let 
\[
k^\ast(i)=\arg\max_{k\neq j(i)} \phi(x_i)_k,
\quad 
u_i=e_{j(i)}-e_{k^\ast(i)}.
\] 
where $e_k$ denotes the $k$-th canonical basis vector in $\mathbb{R}^m$.
The margin condition
$\phi(x_i)_{j(i)}-\max_{k\neq j(i)}\phi(x_i)_k\ \ge\ \gamma
$ is equivalent to $u_i^\top W f(x_i)\ \ge\ \gamma.$

The operator-norm inequality yields
\[
u_i^\top W f(x_i)
\le
\|u_i\|_2\,\|W\|_2\,\|f(x_i)\|_2.
\]
Since $u_i=e_{j(i)}-e_{k^\ast(i)}$ has two nonzero entries $\pm1$,
we have $\|u_i\|_2=\sqrt{2}$.  
Thus for every event $i$,
\[
\gamma
\le 
\sqrt{2}\,\|W\|_2\,\|f(x_i)\|_2.
\]
As a result, we conclude that
\[
\|W\|_2
\ \geq\
\frac{\gamma}{\sqrt{2}\max_i\|f(x_i)\|_2},
\]
which is the desired lower bound. 
\end{proof}

\subsection{Proof of Theorem~\ref{thm:pch-fg}~(Absence of finite-norm interpolation in PC-Hazard)}
\label{app:pch-fg}

\begin{proof}

\noindent Let $\eta_{i,k}=\log(1+e^{z_{\theta,k}(x_i)})>0.$
The per-sample loss is
$\ell_i(\theta)
= -\delta_i\log \eta_{i,j(i)} + \rho_i \eta_{i,j(i)} 
+ \sum_{k<j(i)} \eta_{i,k}.
$
Its infimum is
\[
\mathcal L^\ast
=\sum_{\delta_i=1}\big(1+\log\rho_i\big),
\]
attained by constructing $\eta_{i,j(i)}=1/\rho_i$ for events and $\eta_{i,k}\to 0^+$ for $k<j(i)$. To apply Lemma \ref{lemma:margin-budget-interval}, we need the following three steps.

\noindent\textbf{Step 1: Near-optimality forces small pre-event $\eta$ and event $\eta$ bounded away from $0$.} Assume
\[
\mathcal L_{\mathrm{PCH}}(\theta_\varepsilon)-\mathcal L^\ast \le \varepsilon.
\]
Since the pre-event terms $\eta_{i,k}\ge 0$ contribute $0$ at the optimum, for every $i$
\begin{equation}
\sum_{k<j(i)} \eta_{i,k} \le \varepsilon
\Longrightarrow 
\,\ \eta_{i,k}\le \varepsilon,~\forall k<j(i).
\label{eqn:pch-pre-small}
\end{equation}
For the event term, $g_i(\eta_{i, j(i)})=-\log\eta_{i, j(i)}+\rho_i\eta_{i, j(i)}$ is strictly convex with
unique minimizer $1/\rho_i$ and $g(\eta_{i, j(i)})\to\infty$ as $\eta_{i, j(i)}\downarrow 0$.
Hence for any $c>0$, there exists $c_i>0$ such that
\[
g_i(\eta_{i, j(i)})\le g_i(1/\rho_i)+c \Longrightarrow c_{i} \le \eta_{i, j(i)}.
\]
Thus for all sufficiently small $c$,
\begin{equation}
c_{i} \le \eta_{i, j(i)}
\Longrightarrow
z_{\theta,j(i)}(x_i)
\ge \log(e^{c_{i} }-1)=:C_i.
\label{eqn:pch-event-lower}
\end{equation}

 \noindent\textbf{Step 2: Small pre-event $\eta$ implies very negative logits and a diverging logit gap.} 
By \cref{eqn:pch-pre-small}, when $\varepsilon \le log(2)$, $$\eta_{i,k}=\log(1+e^{z_{\theta,k}(x_i)})\leq \varepsilon\Longrightarrow z_{\theta,k}(x_i) \le \log(e^\varepsilon-1)\le \log\varepsilon.$$

Combining with \cref{eqn:pch-event-lower},
\[
z_{\theta,j(i)}(x_i)-z_{\theta,k}(x_i)
\ \ge\ 
C_i - \log\varepsilon
:=\gamma_i(\varepsilon).
\]
Let $\gamma(\varepsilon)=\min_i \gamma_i(\varepsilon).$
Then $\gamma(\varepsilon)\to\infty$ as $\varepsilon\downarrow 0$.

\noindent\textbf{Step 3: Shared embedding forces $\|W\|\to\infty$.} Write the last layer as
$z_\theta(x)=W f(x)+b, a_i=f(x_i)$.
Let $u_i=e_{j(i)}-e_{k^*(i)}$. The margin can be witten as,
$u_i^\top W a_i \ \ge\ \gamma(\varepsilon)$.

By Lemma \ref{lemma:margin-budget-interval},
\[
\|W\|_2
\ \geq\
\frac{\gamma(\varepsilon)}{\sqrt{2}\max_i \|a_i\|_2}.
\]
Since $\gamma(\varepsilon)\to\infty$, we have $\|W\|_2\to\infty$ whenever
$\mathcal L_{\mathrm{PCH}}(\theta_\varepsilon)\to\mathcal L^\ast$.
\end{proof}

\subsection{Proof of Theorem~\ref{thm:nnet-interp}~(Existence of interpolation in Nnet-Survival)}
\label{app:nnet-interp}

\begin{proof}
We show that Nnet-Survival
admits an interpolation through constructing a finite-dimensional solution by working directly with logits $z_{\theta,k}(x_i)$, in three steps.

\noindent\textbf{Step 1: Per-cell loss.}
For each training cell $(i,j)$ with binary label $y_{ij}\in\{0,1\}$, define the logistic loss
\[
\ell_{ij}(z_{\theta,j}(x_i)) = -\big[y_{ij}\log\sigma(z_{\theta,j}(x_i)) + (1-y_{ij})\log(1-\sigma(z_{\theta,j}(x_i)))\big],
\quad
\sigma(z) = \frac{1}{1+e^{-z}}.
\]
The total loss is
$\mathcal L_{\mathrm{Nnet-Survival}}(\theta) = \sum_{i,j}\ell_{ij}(z_{\theta,j}(x_i))$.

\noindent\textbf{Step 2: Constructing a dominant sequence.}
For any scalar $t>0$, define logits
\[
z_{\theta,j}(x_i)^{(t)} =
\begin{cases}
t, & y_{ij}=1,\\[0.2em]
-t, & y_{ij}=0.
\end{cases}
\]
When $y_{ij}=1$, $\ell_{ij}(z_{\theta,j}(x_i)^{(t)})=-\log\sigma(t)=\log(1+e^{-t})\le e^{-t}$.
When $y_{ij}=0$, $\ell_{ij}(z_{\theta,j}(x_i)^{(t)})=-\log(1-\sigma(-t))=-\log\sigma(t)=\log(1+e^{-t})\le e^{-t}$.
Hence in all cases $\ell_{ij}(z_{\theta,j}(x_i)^{(t)})\le e^{-t}$.\\
\textbf{Step 3: Loss convergence.}
Let $d=mn$ denote the total number of logits $z_{\theta,j}(x_i)$. Because each logit is parametrized by $\theta$, satisfying $d$ logit assignments only requires a parameter vector with at least $d$ degrees of freedom. Then
\[
0 \le \mathcal L_{\mathrm{Nnet-Survival}}(\theta) = \sum_{i,j}\ell_{ij}(z_{\theta,j}(x_i)^{(t)}) \le d e^{-t} \;\xrightarrow[]{t\to\infty}\; 0.
\]
Therefore $\inf_{\theta\in\mathbb{R}^d} \mathcal L_{\mathrm{Nnet-Survival}}(\theta)=0$, and hence $d^*\leq d$ and Nnet-Survival admits an interpolation.
\end{proof}

\subsection{Proof of Theorem~\ref{thm:nnet-no-fg}~(Absence of finite-norm interpolation in Nnet-Survival)}
\label{app:nnet-no-fg}

\begin{proof}
We show the absence of finite-norm interpolation in Nnet-Survival in three steps.

\noindent\textbf{Step 1: Small logistic loss forces polarized logits.}
Consider any cell $(i,j)$ with label $y_{ij}\in\{0,1\}$ and logit $z_{\theta,j}(x_i).$
The per-cell logistic loss is
\[
\ell_{ij}(\theta)
= -\Big[y_{ij}\log\sigma\!\big(z_{\theta,j}(x_i)\big)
   + (1-y_{ij})\log\!\big(1-\sigma(z_{\theta,j}(x_i))\big)\Big].
\]
Suppose $\ell_{ij}(\theta)\le\varepsilon$. If $y_{ij}=1$, then $-\log\sigma(z_{\theta,j}(x_i))\le\varepsilon$, which implies
\[
z_{\theta,j}(x_i) \;\ge\; 
\alpha(\varepsilon)
\;:=\;
\log\!\frac{1}{e^{\varepsilon}-1}.
\]
If $y_{ij}=0$, the loss is symmetric in $z\mapsto -z$, so
$z_{\theta,j}(x_i) \;\le\; -\alpha(\varepsilon)$.

\noindent\textbf{Step 2: Event subjects yield diverging within-subject logit margins.}
If subject $i$ experiences an event in interval $j(i)$, then $y_{i,j(i)}=1$ and
$y_{ik}=0$ for all $k\neq j(i)$.  
By Step~1,
\[
z_{\theta,j(i)}(x_i) \;\ge\; \alpha(\varepsilon),
\quad
z_{\theta,k}(x_i) \;\le\; -\alpha(\varepsilon)
\quad (k\neq j(i)).
\]
Observe that for $0<\varepsilon\le \log 2$,
\[
\alpha(\varepsilon) \;\ge\; \log(1/\varepsilon)-\log 2. 
\]
Thus the within-subject margin satisfies
\[
m_i
:= 
z_{\theta,j(i)}(x_i)
\;-\;
\max_{k\neq j(i)} z_{\theta,k}(x_i)
\;\ge\;
2\,\alpha(\varepsilon)
\;\ge\;
2\log(1/\varepsilon)-2\log 2.
\]
Let $\gamma(\varepsilon)=\min_i m_i$.  
Then $\gamma(\varepsilon)\to\infty$ as $\varepsilon\downarrow 0$.

\noindent\textbf{Step 3: Shared embedding forces weight-norm blow-up.}
From Step~2, each event subject $i$ satisfies a within-subject logit gap 
$m_i \ge \gamma(\varepsilon)\xrightarrow[]{\varepsilon\to0}\infty$. Applying Lemma~\ref{lemma:margin-budget-interval} yields the operator–norm lower bound
\[
\|W\|_2 
\;\geq\;
\frac{\gamma(\varepsilon)}{\sqrt{2}\max_i \| f(x_i)\|_2}\xrightarrow[]{\varepsilon\to0}\infty.
\]
\end{proof}

\subsection{Proof of Theorem~\ref{thm:mtlr-interp}~(Existence of interpolation in N-MTLR)}
\label{app:mtlr-interp}

\begin{proof}
We show that N-MTLR loss admits an interpolation by constructing a
sequence of cumulative logits for which all per-sample loss terms converge to $0$, in three steps.

\noindent\textbf{Step 1: Event samples.}
Suppose subject $i$ experiences an event in interval $j(i)$.  Its loss contribution is
\[
-\log p_{\theta,j(i)}(x_i),
\quad
p_{\theta,j(i)}(x_i)
=
\frac{\exp\!\big(\sum_{k=j(i)}^m z_{\theta,k}(x_i)\big)}
     {\sum_{\ell=1}^m \exp\!\big(\sum_{k=\ell}^m z_{\theta,k}(x_i)\big)}.
\]
Define the cumulative logits for sample $i$ as $C_{\theta,j(i)}(x_i) =\sum_{k=j(i)}^m z_{\theta,k}(x_i)$. Then $p_{\theta,j(i)}(x_i)
= \frac{\exp\!\big(C_{\theta,j(i)}(x_i)\big)}
     {\sum_{\ell=1}^m \exp\!\big(C_{\theta,l}(x_i)\big)}$. Define 
\[
C_{\theta,j(i)}(x_i)^{(t)} =t,
\quad
C_{\theta,j \ne j(i)}(x_i)^{(t)} =-t
\] for any $t>0$. 
Then define the corresponding $p_{\theta,j(i)}(x_i)^{(t)}
=\frac{\exp\!\big(C_{\theta^{(t)},j(i)}(x_i)\big)}
     {\sum_{\ell=1}^m \exp\!\big(C_{\theta^{(t)},l}(x_i)\big)}$. 
\[
p_{\theta,j(i)}(x_i)^{(t)}
=
\frac{e^{t}}
     {e^{t}+\sum_{\ell\neq j(i)}e^{-t}}
\;\ge\;
\frac{e^{t}}{e^{t}+(m-1)e^{-t}}
\;\xrightarrow[]{t\to\infty}\;1,
\]
so $-\log p_{\theta,j(i)}(x_i)^{(t)}\xrightarrow[]{t\to\infty}0$.

\noindent\textbf{Step 2: Censored samples.}
If subject $i$ is censored at interval $j(i)$, its loss term is
\[
-\log S_{\theta,j(i)}(x_i),
\quad
S_{\theta,j(i)}(x_i)=\sum_{\ell=j(i)+1}^m p_{\theta,\ell}(x_i).
\]
For any $t>0$, define construct cumulative logits as
\[
C_{\theta,j \le j(i)}(x_i)^{(t)} = \sum_{k\le j(i)}^m z_{\theta,k}(x_i)^{(t)}=-t,
\quad
C_{\theta,j > j(i)}(x_i)^{(t)} = \sum_{k>j(i)}^m z_{\theta,k}(x_i)^{(t)}=t.
\]
Then
\[
S_{\theta,j(i)}(x_i)^{(t)}
=
\frac{\sum_{\ell>j(i)}e^t}
     {\sum_{\ell>j(i)}e^t+\sum_{\ell\le j(i)}e^{-t}}
\;\ge\;
\frac{(m-j(i))e^t}{(m-j(i))e^t+j(i)e^{-t}}
\;\xrightarrow[]{t\to\infty}\;1,
\]
and thus $-\log S_{\theta,j(i)}(x_i)^{(t)}\xrightarrow[]{t\to\infty} 0$.

\noindent\textbf{Step 3: Realizability through finite many base logits $z_{\theta,k}(x_i)$.}
Based on Steps~1-2, for every sample $i$ and every interval $j$, a target cumulative logit $C_{\theta,j}(x_i)^{(t)}\in\{\,t,\,-t\,\}$. Hence we obtain exactly $n\times m$ cumulative-logit constraints such that both $-\log p_{\theta,j(i)}(x_i)^{(t)}\to0$ and $-\log S_{\theta,j(i)}(x_i)^{(t)}\to 0$.

Because only the cumulative sums 
$C_{\theta,j}(x_i)^{(t)}$ are constrained,
and no additional structural restriction is imposed on the $n\times m$ logits $ z_{\theta,k}(x_i)^{(t)}$, one can always select base logits whose running sums match
$C_{\theta,j}(x_i)^{(t)}$. For example, assigning $z_{\theta,j(i)}(x_i)^{(t)} = C_{\theta,j(i)}(x_i)^{(t)} = t $ and  $z_{\theta,k}(x_i)^{(t)}=0, k \ne j(i)$ realizes the event-sample constraints,
and an analogous assignment realizes the censored-sample constraints.

In particular, if one assigns an independent free parameter to every logit
$z_{\theta,k}(x_i)$, the dimension of the parameter space is at most $mn$. Our construction in Steps~1-2 only constrains the cumulative logits and therefore requires no more than this upper bound.  Consequently, $d^{*}\le d=mn$ so N-MTLR loss admits an interpolation.

\end{proof}

\subsection{Proof of Theorem~\ref{thm:mtlr-no-fg}~(Absence of Finite-Norm Interpolation in N-MTLR)}
\label{app:mtlr-no-fg}

\begin{proof}
For each subject $i$, the model produces base logits
$z_{\theta,1}(x_i),\dots,z_{\theta,m}(x_i)$,
and defines cumulative logits
$C_{\theta,j}(x_i)=\sum_{k=j}^m z_{\theta,k}(x_i), j=1,\dots,m$.
Assume a shared-embedding last layer for the base logits:
$z_{\theta,k}(x)=(W f(x))_k + b_k, f(x)\in\mathbb{R}^u$.
Suppose there exists an $\varepsilon$-interpolating parameter $\theta_\varepsilon$ such that
\[
-\log p_{\theta,j(i)}(x_i)\le\varepsilon
\quad (\delta_i=1),\quad
-\log S_{\theta,j(i)}(x_i)\le\varepsilon
\quad (\delta_i=0),
\]
for all subjects $i$, where $p_{\theta,j}(x_i)$ and $S_{\theta,j(i)}(x_i)$ are the MTLR event and survival probabilities. We show the absence of finite-norm interpolation of N-MTLR in three steps.

\noindent\textbf{Step 1: Small loss forces large cumulative-logit margins.}
Consider first an event subject $i$ with $\delta_i=1$ and event interval $j(i)$.  
The loss condition $-\log p_{\theta,j(i)}(x_i)\le\varepsilon$ implies
\[
p_{\theta,j(i)}(x_i)\ \ge\ e^{-\varepsilon}.
\]
Since $p_{\theta,j}(x_i)$ is a softmax in the cumulative logits $\{C_{\theta,\ell}(x_i)\}_{\ell=1}^m$, standard softmax tail bounds yield the existence of constants $c_1,c_0>0$ which are independent of $\varepsilon$, such that for all sufficiently small $\varepsilon$,
\[
C_{\theta,j(i)}(x_i)
- \max_{\ell\neq j(i)} C_{\theta,\ell}(x_i)
\ \ge\ c_1\log(1/\varepsilon)-c_0
\ \gtrsim\ \log(1/\varepsilon),
\]
where $a \gtrsim b$ means $a \ge c \times b - O(1)$ for some constant $0<c<\infty$.

For censored subjects ($\delta_i=0$), the condition $-\log S_{\theta,j(i)}(x_i)\le\varepsilon$ implies $S_{\theta,j(i)}(x_i)\ge e^{-\varepsilon}$, i.e., the survival probability after $j(i)$ is close to one. An analogous softmax argument applied to the tail probabilities $\{p_{\theta,\ell}(x_i)\}_{\ell>j(i)}$ yields a cumulative-logit gap of the same order, again $\gtrsim\log(1/\varepsilon)$.  

In either case, for every subject $i$ there exist indices $\tilde{k}$ and $\tilde{l}$ such that
\begin{equation}
C_{\theta,\tilde{k}}(x_i)-C_{\theta,\tilde{l}}(x_i)
\ \gtrsim\ \log(1/\varepsilon).
\label{eqn:mtlr-cum-gap}
\end{equation}

\noindent\textbf{Step 2: Cumulative-logit gaps force large base-logit norms.}
For a fixed subject $i$, let $C_i\in\mathbb{R}^m$ denote the vector of cumulative logits $C_{\theta,\bullet}(x_i)$.  
From \cref{eqn:mtlr-cum-gap} and the fact that the difference between any two coordinates of a vector is at most $\sqrt{2}\,\|C_i\|_2$, we obtain
$\|C_i\|_2\ \gtrsim\ \log(1/\varepsilon)$.

Next, relate the base-logit vector $z_i=(z_{\theta,1}(x_i),\dots,z_{\theta,m}(x_i))^\top$ to $C_i$.  
By definition,
$C_{\theta,j}(x_i)=\sum_{k=j}^m z_{\theta,k}(x_i)$,
so each coordinate $C_{\theta,j}(x_i)$ is a sum of at most $m$ base logits.  
Hence, by Cauchy-Schwarz inequality,
\[
|C_{\theta,j}(x_i)|
\ \le\ \sum_{k=j}^m |z_{\theta,k}(x_i)|
\ \le\ \sqrt{m}\,\|z_i\|_2,
\]
and therefore
\[
\|C_i\|_2
\ =\ \Big(\sum_{j=1}^m C_{\theta,j}(x_i)^2\Big)^{1/2}
\ \le\ \sqrt{m}\,\max_j |C_{\theta,j}(x_i)|
\ \le\ m\,\|z_i\|_2.
\]
Combining this with the lower bound on $\|C_i\|_2$ gives
$\|z_i\|_2\ \gtrsim\ \frac{\log(1/\varepsilon)}{m}$.
In particular, for each $i$ there exists some coordinate $k(i)$ such that
\begin{equation}
|z_{\theta,k(i)}(x_i)|
\ \gtrsim\ \frac{\log(1/\varepsilon)}{m}.
\label{eqn:mtlr-large-z}
\end{equation}

\noindent\textbf{Step 3: Shared embedding forces weight-norm blow-up.}
Under the shared-embedding parameterization
\[
z_{\theta,k}(x) = (W f(x))_k + b_k,
\]
this implies that at least one coordinate of $W f(x_i)$ must grow on the same order as $\log(1/\varepsilon)$.  
By Lemma~\ref{lemma:margin-budget-interval}, this two-coordinate comparison yields
\[
\|W\|_2
\;\gtrsim\;
\frac{\log(1/\varepsilon)}{\max_i \|f(x_i)\|_2}\xrightarrow[]{\varepsilon\to0}\infty.
\]
Thus N-MTLR does not admit a finite-norm interpolation.
\end{proof}
\hfill$\blacksquare$

\section{Additional experimental details}
All four neural-network models were trained using the Adam optimizer with its standard default momentum parameters. We evaluated combinations of batch size $\{32, 64, 128, 256\}$ and learning rate $\{5\times 10^{-5},\,1\times 10^{-4},\,3\times 10^{-4},\,5\times 10^{-4},\,10^{-3},\,2\times 10^{-3}\ , 10^{-2}\}$.
For each combination, we trained a two-layer network for $30$ independent replicates and terminated training once the training loss no longer exhibited substantial  decrease. We recorded both the training and test losses at
convergence, averaged these over all replicates, and reported the median across hyperparameter combinations for each network width.

For the discretized-time models, PC-Hazard used $50$ time intervals, whereas
Nnet-Survival and N-MTLR used $20$ intervals; these choices follow common
practice in survival neural networks and provide sufficiently fine
approximations for the datasets considered.

Code for all simulations can be found in the GitHub repository: \url{https://github.com/Maeve816/double–-descent-survival}.

\vskip 0.2in
\bibliographystyle{plainnat}
\bibliography{refs}

\end{document}